\documentclass{article}
\usepackage{ijcai26}

\usepackage{times}
\usepackage{soul}
\usepackage{url}
\usepackage[hidelinks]{hyperref}
\usepackage[utf8]{inputenc}
\usepackage[small]{caption}
\usepackage{graphicx}
\usepackage{amsmath}
\usepackage{amsthm}
\usepackage{tabularx}
\usepackage{booktabs}
\usepackage{algorithm}
\usepackage{algorithmic}
\usepackage[switch]{lineno}
\usepackage{graphicx}
\usepackage{multirow} 
\usepackage{makecell}
\usepackage{CJKutf8}
\usepackage[most]{tcolorbox}
\usepackage{cleveref}
\usepackage{subcaption}
\usepackage[table]{xcolor}
\usepackage{colortbl}

\title{Probing Cultural Awareness in LLMs: A Case Study of Cross-Culture \\ Aesthetic Stylistics}

\author{
Jiashuo Wang$^1$\thanks{Equal contribution.}\and
Fenggang Yu$^1$\footnotemark[1]\and
Jian Wang$^1$\and
Chak Tou Leong$^1$\and
Xiaoyu Shen$^2$\and\\
Chunpu Xu$^1$\and
Jiawen Duan$^3$\and
Wenjie Li$^1$\and
Johan F. Hoorn$^{1,4,5,6}$
\affiliations
$^{1}$ Department of Computing, Hong Kong Polytechnic University\\
$^{2}$ Institute of Digital Twin, Eastern Institute of Technology, Ningbo \\
$^{3}$ Department of Language Science and Technology, Hong Kong Polytechnic University \\
$^{4}$ School of Design, Hong Kong Polytechnic University\\
$^{5}$ Research Institute for Quantum Technology, Hong Kong Polytechnic University\\
$^{6}$ Department of Communication Science, Vrije Universiteit Amsterdam\\
\emails
\texttt{\{jessie25.wang,jian51.wang,wenjie.li,johan.f.hoorn\}@polyu.edu.hk}\\\texttt{\{fenggang.yu,chun-pu.xu,jiawen.duan\}@connect.polyu.hk}
}

\begin{document}

\maketitle

\begin{abstract}
Large Language Models (LLMs) are increasingly deployed in diverse cultural contexts, yet their ability to master aesthetic stylistics, i.e., the strategic use of language to evoke cultural resonance, remains underexplored. We curate \textbf{\textsc{C\textsuperscript{4}Styli}}, a benchmark of highly stylized translated movie titles and advertising slogans from Hong Kong and the Chinese Mainland, to evaluate LLMs via the lens of behavioral recognition and productive competence. Extensive evaluations show that LLMs differ from humans in stylistic recognition, and this recognition ability varies across text domains. In addition, stylistic recognition and generation performance in LLMs are not consistently aligned.
To further examine whether LLMs genuinely capture stylistic information in stylistic recognition, we conduct structural ablation with logistic regression probes. We find that, in the Hong Kong setting, stylistic recognition in LLMs relies primarily on surface-level linguistic information rather than stylistic structure. This suggests limited sensitivity to Hong Kong-specific stylistic structure. Our code and data are available at \url{https://github.com/wangjs9/C4STYLI}.
\end{abstract}

\section{Introduction}
Recent advances in Large Language Models (LLMs) have demonstrated remarkable capabilities in tasks such as question answering \cite{kim2024learning}, reasoning \cite{zhangllm}, and dialogue \cite{wang2024instruct}. Consequently, studying their cultural awareness, the ability to recognize and appropriately respond to cultural nuances \cite{pawar2025survey}, has become critical for developing trustworthy, culturally aligned applications across diverse populations.

\begin{figure}[t]
    \centering
    \includegraphics[width=.85\linewidth]{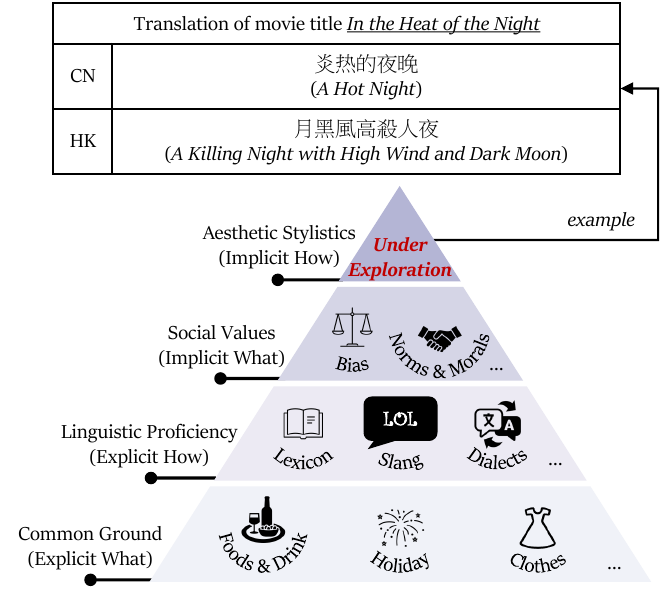}
    \caption{Hierarchy of cultural awareness in LLMs.}
    \label{fig:pyramid}
\end{figure}

Culture encompasses various dimensions, ranging from tangible artifacts to intangible values. Among these, \textbf{aesthetic stylistics} stands as one of the most sophisticated yet under-explored facets in LLM research. It represents the strategic choice made among various grammatically correct and semantically equivalent ways of expressing the same content. \cite{hickey2014pragmatics}. Unlike explicit facts or linguistic rules (\Cref{fig:pyramid}), we conceptualize aesthetic stylistics as \textit{an implicit way that determines how to tailor the expression of specific content} to attract the reader's attention and provoke resonance without altering the core semantics \cite{riffaterre1978semiotics}.

\begin{figure*}[t!]
    \centering
    \includegraphics[width=\linewidth]{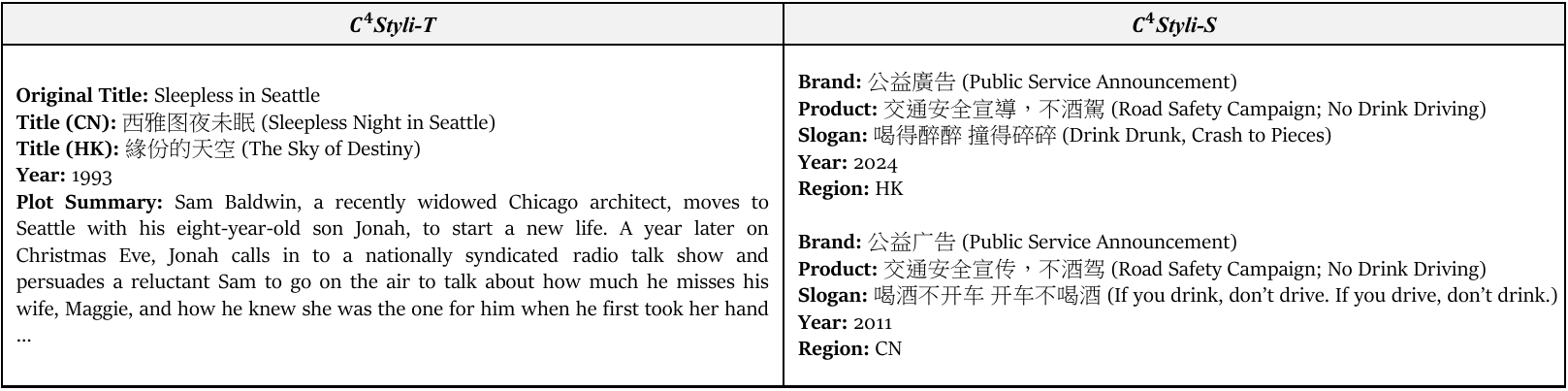}
    \caption{An instance from \textsc{C\textsuperscript{4}Styli}-T (left) and two instances from \textsc{C\textsuperscript{4}Styli}-S (right). English translations are provided for illustration.}
    \label{fig:instance}
\end{figure*}

This work investigates cultural awareness in LLMs through aesthetic stylistics. We introduce \textbf{\textsc{C\textsuperscript{4}Styli}} ($C$ross-$C$ulture $C$hinese-$C$hinese \textbf{Styli}stics), a curated benchmark including translated movie titles and advertising slogans in the Chinese Mainland (CN) and Hong Kong (HK). This dataset offers two primary advantages. (1) \textbf{Minimal Linguistic Interference}: By utilizing the mutual convertibility between Simplified and Traditional Chinese, we minimize surface-level linguistic variance, allowing for a more focused analysis of cultural aesthetic stylistics. (2) \textbf{High Stylistic Salience}: Unlike neutral prose, these texts are purposely engineered to maximize resonance and attention, providing a high-signal environment for observing aesthetic stylistics.

With this dataset, we analyze current popular LLMs' capacities in terms of their behavioral recognition, productive competence, and internal representation. Our analysis yields three primary findings: \textbf{(1)} LLMs show clear differences from humans in recognizing cultural stylistics, and their performance varies substantially across different text domains. \textbf{(2)} We observe a decoupling between recognition and production: models that excel at identifying styles do not necessarily mirror that proficiency in generating them. \textbf{(3)} Through structural ablation, we reveal that LLMs' awareness of HK culture remains superficial. While models can ``match'' HK styles through sparse, high-intensity lexical anchors, they lack the integrated structural understanding observed in their representation of CN styles.

Our contributions are as follows: \textbf{(1) Benchmark:} We introduce \textbf{\textsc{C\textsuperscript{4}Styli}}, which mitigates surface linguistic interference via script normalization to focus on implicit aesthetic stylistics. \textbf{(2) Behavioral Gap:} We find LLM–human differences, domain variation, and a decoupling between stylistic recognition and production. \textbf{(3) Shallow Encoding of HK Stylistics:} We find that LLMs encode HK style via superficial lexical shortcuts instead of integrated structures.

\section{Related Work}
\paragraph{Cultural Awareness in LLMs.}
Culture is an expansive and multifaceted construct, necessitating a granular approach to its study within LLMs. To systematically examine cultural awareness, we organize its various dimensions into a hierarchical pyramid structure (\Cref{fig:pyramid}), reflecting a progression from explicit manifestations to implicit nuances. At the base of our framework lie the foundational layers, representing the explicit ``what'' and ``how'' of culture. \textit{Common Ground} encompasses tangible cultural artifacts and factual concepts, such as traditional clothing \cite{zhou-etal-2025-hanfu} and regional cuisine \cite{zhou-etal-2025-mapo}. \textit{Linguistic Proficiency} focuses on the formal communicative conventions and regional codes, including specialized lexicons \cite{cao2024cultural}, slang \cite{sun-xu-2022-tracing}, and regional dialects \cite{faisal-anastasopoulos-2025-testing}. The higher echelons of the pyramid transition into the implicit realm. \textit{Social Values} represent the implicit ``what'', governing how language adheres to societal morals and ethics to ensure appropriateness within cultural norms. Prior work in this layer has focused on identifying cultural biases \cite{nangia-etal-2020-crows}, aligning models with regional norms and morals \cite{li2024culturellm}, and evaluating cross-cultural ethical reasoning \cite{mohammadi2025large}. Finally, at the apex sits \textit{Aesthetic Stylistics}, which we define as the implicit ``how.'' This sophisticated layer dictates the artistic treatment of language required to achieve pragmatic effects and stimulate profound resonance within an audience.

\paragraph{Cantonese NLP.}
Cantonese (Yue), despite being used by a large and geographically dispersed speech community, remains comparatively underrepresented in current NLP research. Even for classical NLP tasks, such as sentiment classification \cite{xiang2019sentiment}, rumor detection \cite{chen2020novel}, and dialogue generation \cite{10.1016/j.csl.2024.101637}, the number of Cantonese-focused studies remains limited.
Unlike simplified Chinese (Zh) and English (En), for which a wide range of mature processing tools are available (e.g., cntext, NLTK, SpaCy), Cantonese lacks comparable tool support, with only a small number of dedicated resources such as PyCantonese \cite{lee2022pycantonese}. This scarcity of linguistic infrastructure further constrains large-scale empirical research on Cantonese.
In the era of LLMs, Cantonese-specific models are still rare. At present, several models released by SenseTime constitute the few publicly available non-commercial LLMs with explicit Cantonese support. Correspondingly, evaluation benchmarks for Cantonese remain limited. Earlier resources were primarily constructed from television programs for machine translation \cite{lee2011toward} or for linguistic annotation \cite{leung2001hkcac}. More recently, a small number of benchmarks, such as Yue-GSM8K, Yue-MMLU, and Yue-ARC, have been proposed to evaluate the reasoning abilities of both Cantonese-specific and general-purpose LLMs \cite{jiang2025well}.
\section{Dataset: \textsc{C\textsuperscript{4}Styli}}
\subsection{Overview}
\textsc{C\textsuperscript{4}Styli} consists of two subcomponents. \textbf{\textsc{C\textsuperscript{4}Styli}-T} contains Chinese movie titles in HK and CN, translated from English source titles. \textbf{\textsc{C\textsuperscript{4}Styli}-S} comprises advertisement slogans in HK and CN. \Cref{fig:instance} shows instances. The dataset focuses on texts in domains that can provide classic instances for studying language style \cite{wales2014dictionary}. Accordingly, \textsc{C\textsuperscript{4}Styli} serves as a diagnostic testbed for analyzing the extent to which cultural stylistic preferences in HK and CN are distinguishable at the level of highly stylized short texts. 

\subsection{Collection and Annotation}
\paragraph{\textsc{C\textsuperscript{4}Styli}-T.}
Movie titles were collected from TMDb (The Movie Database)
and Wikipedia. Using a web-scraping framework, we extracted the titles and plot summaries in English, along with their officially released Chinese titles for both HK and CN. Instances were manually filtered if they met any of the following criteria: (1) the Chinese title matches the original English title, (2) the plot summary is missing, or (3) the Chinese titles in CN and HK are identical (ignoring simplified vs. traditional script). Each instance includes the English title, the movie release year, the plot summary in English, and translated titles in HK and CN.

\begin{table}[tb]
\centering
\small
\resizebox{\linewidth}{!}{\begin{tabular}{ccccc}
\toprule
\textbf{Subsplit} & \textbf{\# Instances} & \textbf{\# HK : \# CN} & \textbf{Avg. Len} & \textbf{Year Range}\\
\midrule
\textsc{C\textsuperscript{4}Styli}-T & 1,530 & 1,530~:~1,530 & 5.10 & 1907-2027 \\
\textsc{C\textsuperscript{4}Styli}-S & 2,837 & 1,976~:~~~~861 & 47.50 & 1953-2025\\
\bottomrule
\end{tabular}}
\caption{Statistics of the \textsc{C\textsuperscript{4}Styli} dataset.}
\label{tab:dataset_stats}
\end{table}

\paragraph{\textsc{C\textsuperscript{4}Styli}-S.}
Advertising slogans were collected from company websites, social media platforms, Bilibili, and YouTube. For content from company websites and social media, we manually extracted the brand, product, year, region, and slogan information. For Bilibili and YouTube, we first identified seed videos containing the keywords (\begin{CJK}{UTF8}{gbsn}经典广告\end{CJK} or \begin{CJK}{UTF8}{bsmi}經典廣告\end{CJK}) in their titles and then collected their related videos, including recommended videos and videos in the same default playlist. We retained only videos whose titles contained the brand name. Slogans were primarily extracted from available subtitles; if subtitles were missing or incomplete, we used qwen3-asr-flash, a speech-to-text system supporting both Mandarin and Cantonese, to generate them. All slogans were manually verified to ensure quality. Each instance consists of the brand name, product information, the release year, the region, and the slogan.

\Cref{tab:dataset_stats} summarizes the overall statistics of \textsc{C\textsuperscript{4}Styli}, including the instance number, average length of target texts (i.e., translated movie titles and slogans), and temporal coverage. It primarily reflects the differences across text domains.

\begin{figure}[t]
    \centering
    \includegraphics[width=\linewidth]{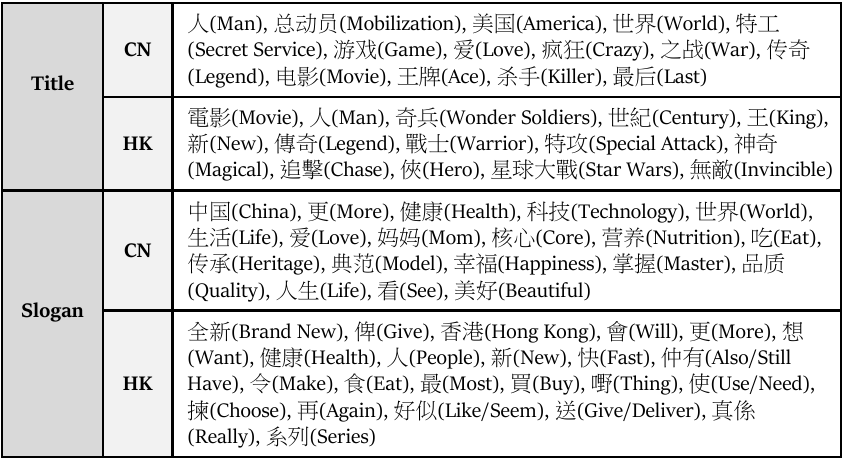}
    \caption{Comparison of the most frequent words used across different text domains and regions.}
    \label{fig:top_words}
\end{figure}

\begin{figure*}[tb]
    \centering
    \begin{subfigure}[t]{0.49\textwidth}
        \centering
        \includegraphics[width=.9\textwidth]{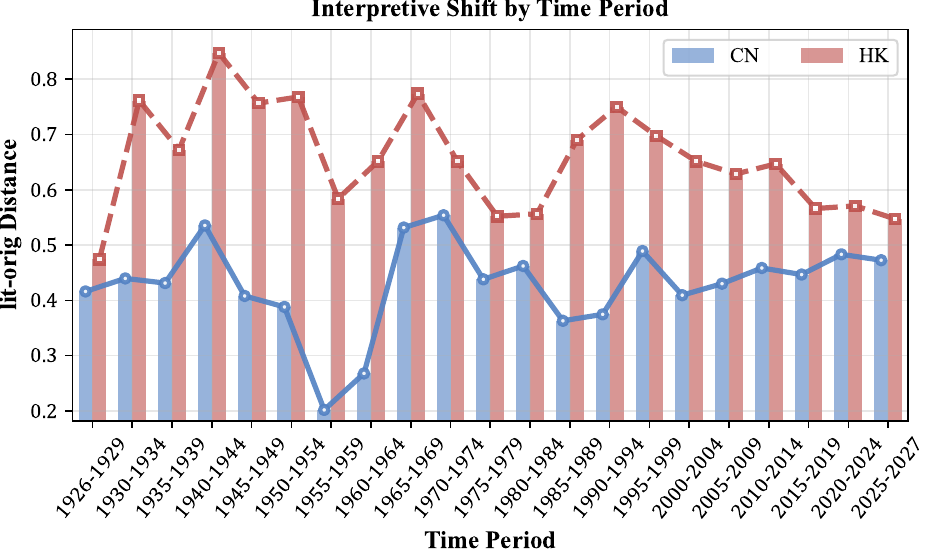}
        \caption{Semantic divergence between translated titles, measured after back-translation into English, and original titles.}
    \end{subfigure}
    \hfill
    \begin{subfigure}[t]{0.49\textwidth}
        \centering
        \includegraphics[width=.9\textwidth]{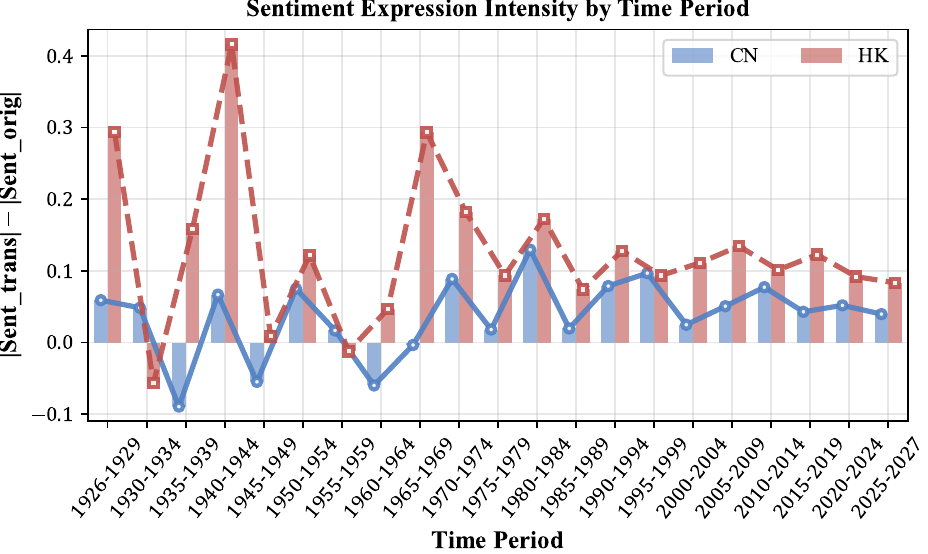}
        \caption{Divergence in affective content between translated titles and original titles, quantified as the difference in sentiment scores.}
    \end{subfigure}
    \caption{Temporal variation in source--target divergence of \textsc{C\textsuperscript{4}Styli}-T in CN and HK.}
    \label{fig:temporal_title}
\end{figure*}

\begin{figure*}[tb]
    \centering
    \begin{subfigure}[t]{0.49\textwidth}
        \centering
        \includegraphics[width=.9\textwidth]{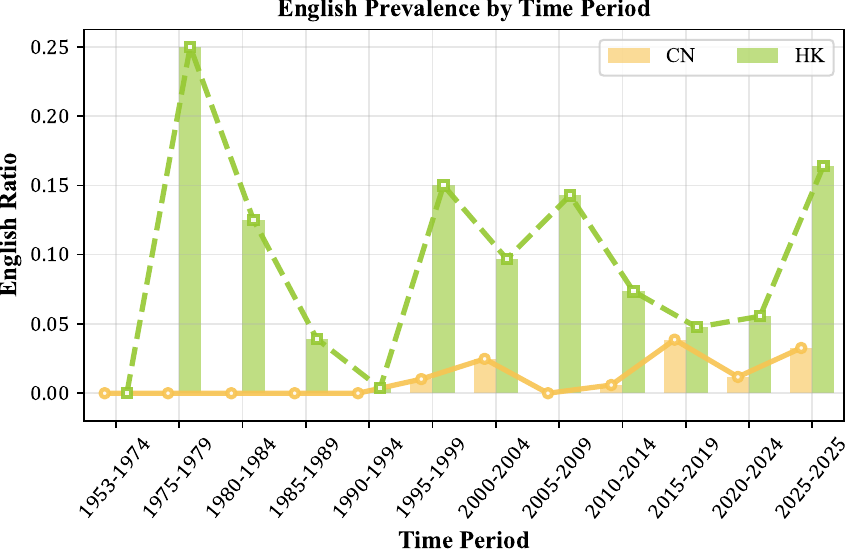}
        \caption{Prevalence rate of English tokens.}
    \end{subfigure}
    \hfill
    \begin{subfigure}[t]{0.49\textwidth}
        \centering
        \includegraphics[width=.9\textwidth]{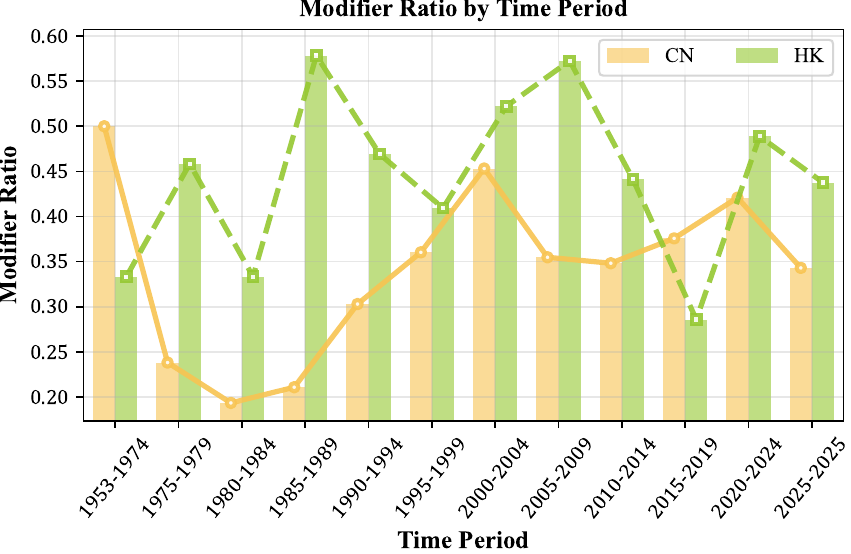}
        \caption{Modifier-to-noun ratio, defined as $\frac{(\# \text{adjectives} + \# \text{adverbs})}{\# \text{nouns}}$.}
    \end{subfigure}
    \caption{Temporal variation of stylistic features in \textsc{C\textsuperscript{4}Styli}-S from CN and HK.}
    \label{fig:temporal_slogan}
\end{figure*}

\subsection{Linguistic Features}
To characterize regional linguistic variation, we analyze differences in language use and lexical preferences between CN and HK. For translated movie titles, we examine their stylistic properties by comparing them with their original English counterparts. Specifically, we quantify source--target divergence by computing $1 - \cos(\mathbf{e}_{bt}, \mathbf{e}_{src})$, where $\mathbf{e}_{bt}$ denotes the embedding of the back-translated Chinese title and $\mathbf{e}_{src}$ denotes the embedding of the original English title, both represented in English embedding space. Our results, HK: 0.617 vs.\ CN: 0.447, show that \textbf{titles released in HK exhibit a larger source--target divergence than those in CN, indicating a greater degree of deviation from the source titles}. 
A top frequency-based lexical analysis further reveals systematic stylistic differences between the two regions, as shown in \Cref{fig:top_words}. CN titles tend to preserve stronger formal equivalence to the source text, frequently relying on literal renderings of high-frequency content nouns (e.g., ``Man,'' ``American,'' ``World''). In contrast, HK titles more often employ expressive reformulations by introducing affectively salient modifiers. 
For example, the film \textit{King Arthur} is rendered in HK as \begin{CJK}{UTF8}{bsmi}王者無敵\end{CJK} (\textit{lit.} King Invincible), where the modifier \begin{CJK}{UTF8}{bsmi}無敵\end{CJK} (Invincible) enhances emotional salience beyond the source title. \textbf{These observations point to distinct stylistic preferences rather than rigid translation conventions across cultures}. To further examine affective differences, we measure the change in sentiment expression between a translated title and its original English version, defined as: $|\text{senti}_{\text{bt}}| - |\text{senti}_{\text{src}}|$, where $\text{senti}$ denotes the sentiment polarity score computed using the VADER analyzer.
We find that HK titles exhibit a larger increase in absolute sentiment strength than CN titles (HK: 0.111 vs.\ CN: 0.052), suggesting that \textbf{HK translations more frequently amplify affective expression relative to the source text than their CN counterparts}.

For advertising slogans, we first examine the prevalence of English usage. 
Specifically, we compute the proportion of English tokens in slogans from each region. 
English accounts for 12.36\% of tokens in HK slogans, compared to only 1.47\% in CN slogans, reflecting \textbf{HK's longstanding bilingual sociolinguistic context}. We further analyze syntactic and lexical stylistic differences by computing the modifier ratio, defined as (\# adjectives + \# adverbs) / \# nouns, for slogans in both regions. The modifier ratio is substantially higher in HK (51.79\%) than in CN (34.67\%), indicating systematic differences in stylistic composition. 
Specifically, \textbf{CN slogans tend to emphasize substantive attributes through noun-heavy constructions, whereas HK slogans more frequently rely on predicate-heavy formulations}. These tendencies are also reflected in the most frequent lexical items shown in \Cref{fig:top_words}. CN slogans often foreground nouns such as \begin{CJK}{UTF8}{gbsn}科技\end{CJK} (Technology) and \begin{CJK}{UTF8}{gbsn}传承\end{CJK} (Heritage), contributing to a more formal and descriptive register. In contrast, HK slogans favor adjectives and adverbs (e.g., \begin{CJK}{UTF8}{bsmi}快\end{CJK} [Fast], \begin{CJK}{UTF8}{bsmi}真係\end{CJK} [Really]) to achieve greater conversational immediacy.

\subsection{Temporal Characteristics}
Having characterized aggregate stylistic differences between cultures, we next examine whether these features remain temporally stable within each culture. 
\Cref{fig:temporal_title} illustrates the temporal variation of source--target divergence and affective change in translated movie titles across cultures. Across all time periods, titles released in HK consistently exhibit larger semantic divergence from their original English counterparts than those released in CN. A similar temporal pattern is observed for affective divergence. As shown in \Cref{fig:temporal_title}(b), HK titles generally demonstrate a larger increase in absolute sentiment strength relative to the original English titles, whereas CN titles exhibit smaller and more conservative changes. 
Notably, although this cultural contrast remains robust throughout the time span, the magnitude of the gap shows signs of narrowing in more recent periods.

\Cref{fig:temporal_slogan} presents the temporal variation of stylistic features in advertising slogans, focusing on English usage and syntactic composition. 
Across all time periods, HK slogans exhibit substantially higher English token ratios than those from CN. While English prevalence in HK slogans fluctuates over time, the overall separation between the two cultures remains evident throughout the entire temporal span. 
In terms of syntactic composition, the modifier ratio also displays consistent cultural differences over time.

From the results, \textit{the cultural differences are temporally ordered rather than merely average-based}: across all observed time periods, the maximum value observed for CN remains below the minimum value observed for HK. In other words, these differences characterize distributional tendencies at the population level, rather than providing sufficient cues for unambiguous culture identification at the level of individual texts or for explicit cultural attribution.

\begin{table*}[t!]
\centering
\small
\resizebox{.95\textwidth}{!}{\begin{tabular}{cccccccccc}
\toprule
\multicolumn{2}{c}{\multirow{3}{*}{\textbf{Model}}} & \multicolumn{4}{c}{\textbf{\textsc{C\textsuperscript{4}Styli}-T}} & \multicolumn{4}{c}{\textbf{\textsc{C\textsuperscript{4}Styli}-S}} \\
\cmidrule(lr){3-6} \cmidrule(lr){7-10}
& & \textbf{Accuracy} & \textbf{Precision} & \makecell[c]{\textbf{Recall}\\(\textbf{macro} \textbf{/} \textbf{HK} \textbf{/} \textbf{CN})} & \textbf{F1} & \textbf{Accuracy} & \textbf{Precision} & \makecell[c]{\textbf{Recall}\\(\textbf{macro} \textbf{/} \textbf{HK} \textbf{/} \textbf{CN})} & \textbf{F1} \\
\midrule
\multirow{7}{*}{En}& o4-mini & \textbf{0.701} & 0.701 & \textbf{0.701} \textbf{/} 0.681 \textbf{/} 0.720 & \textbf{0.701} & \textbf{0.788} & \textbf{0.836} & 0.726 \textbf{/} 0.479 \textbf{/} \textbf{0.973} & 0.740\\
& GPT-4.1 & 0.685  & \textbf{0.705} & 0.685 \textbf{/} 0.529 \textbf{/} 0.841 & 0.677 & 0.819  & 0.828 & \textbf{0.797} \textbf{/} \textbf{0.675} \textbf{/} 0.920 & \textbf{0.806} \\
& Llama-3-8B-Instruct & 0.535  & 0.546 & 0.540 \textbf{/} 0.729 \textbf{/} 0.351 & 0.520 & 0.687  & 0.748 & 0.611 \textbf{/} 0.256 \textbf{/} 0.966 & 0.590 \\
& Llama-3-70B-Instruct & 0.624  & 0.637 & 0.627 \textbf{/} \textbf{0.765} \textbf{/} 0.490 & 0.618 & 0.699  & 0.688 & 0.675 \textbf{/} 0.548 \textbf{/} 0.802 & 0.678 \\
& gemma-3-4b-it & 0.558  & 0.571 & 0.552 \textbf{/} 0.293 \textbf{/} 0.810 & 0.522 & 0.691  & 0.685 & 0.691 \textbf{/} 0.693 \textbf{/} 0.690 & 0.686 \\
& gemma-3-12b-it & 0.547  & 0.713 & 0.537 \textbf{/} 0.083 \textbf{/} \textbf{0.991} & 0.422 & 0.746  & 0.782 & 0.701 \textbf{/} 0.459 \textbf{/} 0.943 & 0.705 \\
& gemma-3-27b-it & 0.604  & 0.626 & 0.608 \textbf{/} 0.800 \textbf{/} 0.416 & 0.591 & 0.779  & 0.797 & 0.746 \textbf{/} 0.567 \textbf{/} 0.925 & 0.754 \\
\midrule
\multirow{9}{*}{Zh} & DeepSeek-V3.2 & 0.700 & 0.719 & 0.700 \textbf{/} 0.553 \textbf{/} \textbf{0.846} & 0.693 & 0.824 & \textbf{0.866} & 0.775 \textbf{/} 0.571 \textbf{/} 0.979 & 0.793\\
& DeepSeek-V3.2 (Thinking) & 0.745  & 0.756 & 0.745 \textbf{/} 0.847 \textbf{/} 0.642 & 0.742 & 0.754  & 0.764 & 0.721 \textbf{/} 0.542 \textbf{/} 0.900 & 0.727 \\
& Qwen2.5-7B-Instruct & 0.598  & 0.618 & 0.593 \textbf{/} 0.368 \textbf{/} 0.819 & 0.574 & 0.761  & 0.760 & 0.737 \textbf{/} 0.608 \textbf{/} 0.865 & 0.743 \\
& Qwen2.5-14B-Instruct & 0.648  & 0.651 & 0.650 \textbf{/} 0.705 \textbf{/} 0.594 & 0.648 & 0.804  & 0.809 & 0.783 \textbf{/} 0.663 \textbf{/} 0.902 & 0.790 \\
& Qwen2.5-32B-Instruct & 0.686  & 0.692 & 0.688 \textbf{/} 0.769 \textbf{/} 0.606 & 0.685 & \textbf{0.831} & 0.829 & \textbf{0.819} \textbf{/} \textbf{0.751} \textbf{/} 0.887 & \textbf{0.823} \\
& Qwen2.5-72B-Instruct & \textbf{0.799}  & \textbf{0.814} & \textbf{0.799} \textbf{/} \textbf{0.910} \textbf{/} 0.687 & \textbf{0.796} & 0.810  & 0.820 & 0.785 \textbf{/} 0.651 \textbf{/} \textbf{0.919} & 0.794 \\
& Qwen3-8B & 0.661  & 0.673 & 0.663 \textbf{/} 0.785 \textbf{/} 0.542 & 0.657 & 0.707  & 0.699 & 0.682 \textbf{/} 0.546 \textbf{/} 0.818 & 0.686 \\
& Qwen3-14B & 0.685  & 0.686 & 0.685 \textbf{/} 0.712 \textbf{/} 0.659 & 0.685 & 0.733  & 0.737 & 0.702 \textbf{/} 0.527 \textbf{/} 0.876 & 0.707 \\
& Qwen3-32B & 0.680  & 0.685 & 0.680 \textbf{/} 0.767 \textbf{/} 0.593 & 0.677 & 0.706  & 0.700 & 0.678 \textbf{/} 0.522 \textbf{/} 0.833 & 0.681 \\
\midrule
\multirow{2}{*}{Yue} & SenseChat-5-Cantonese & 0.623  & 0.631 & 0.623 \textbf{/} \textbf{0.752} \textbf{/} 0.494 & 0.616 & 0.586  & \textbf{0.775} & 0.582 \textbf{/} 0.164 \textbf{/} \textbf{1.000} & 0.495 \\
& SenseNova-V6-5-Turbo & \textbf{0.652} & \textbf{0.654} & \textbf{0.651} \textbf{/} 0.592 \textbf{/} \textbf{0.710} & \textbf{0.650} & \textbf{0.762} & 0.761 & \textbf{0.715} \textbf{/} \textbf{0.528} \textbf{/} 0.901 & \textbf{0.725}\\
\midrule
\multicolumn{2}{c}{Human Baseline} & \textbf{0.897} & \textbf{0.897} & \textbf{0.897} \textbf{/} \textbf{0.901} \textbf{/} \textbf{0.893} & \textbf{0.897} & \textbf{0.933} & \textbf{0.930} & \textbf{0.935} \textbf{/} \textbf{0.948} \textbf{/} \textbf{0.922} & \textbf{0.932}\\
\bottomrule
\end{tabular}}
\caption{LLM performance in cultural stylistic identification across \textsc{C\textsuperscript{4}Styli}-T and \textsc{C\textsuperscript{4}Styli}-S.}
\label{tab:identification}
\end{table*}
\section{Overall Setup}
\paragraph{Assessed LLMs.}
For behavioral recognition and productive competence, we evaluate the cultural awareness of 18 LLMs, spanning six major model families: (1) OpenAI series, (2) Meta-Llama series, (3) Google Gemma series, (4) DeepSeek series, (5) Qwen series, and (6) SenseNova series. These models are categorized according to their relatively stronger languages, i.e., En, Zh, and Yue.

\paragraph{Dataset Splitting.}
We split the dataset into three subsets. For \textsc{C\textsuperscript{4}Styli}-T, we randomly selected 1010 instances for training and 520 for testing. For \textsc{C\textsuperscript{4}Styli}-S, we randomly selected 2,287 instances for training and 550 for testing.

\section{Can LLMs Discriminate Cultural Stylistics?}
\paragraph{Method.}
To investigate whether LLMs can identify culturally grounded stylistic distinctions, we formulate a classification task that evaluates whether LLMs can discriminate cultural stylistics by predicting the cultural origin (HK vs.\ CN) of a given text, conditioned on relevant contextual information.
Specifically, for translated movie titles, we provide the original title and plot summary as context; for advertising slogans, we supply the brand name, product information, and year of release.
We examine model behavior under different prompting conditions by varying (i) the language variety used in the prompt (Cantonese vs.\ Mandarin) and (ii) the presence or absence of in-context exemplars.

\paragraph{Evaluation.}
The performance is evaluated by standard classification metrics, including \textbf{accuracy}, \textbf{precision}, \textbf{recall}, and \textbf{F1 score}.
All metrics are reported in their macro-averaged form to account for potential class imbalance.
In addition to macro recall, we separately report recall scores by treating HK and CN as the positive class, respectively, to assess potential asymmetries in model sensitivity across cultures.

\paragraph{LLM Overall Performance.}
\Cref{tab:identification} summarizes LLM performance on the cultural stylistic identification task. Each model is evaluated using four prompt variants, with the reported results averaged across variants. A human baseline is established by asking two human participants to identify the cultural style class of each text; inter-annotator agreement reaches a Krippendorf's $\alpha$ of 0.613, indicating substantial agreement. Overall, \textbf{LLM performance remains substantially below the human baseline, revealing a persistent gap in cultural stylistic recognition}. Beyond this overall performance difference, the human–model gap is further evidenced along two additional dimensions. First, \textbf{LLMs exhibit strong sensitivity to text type}. Advertising slogans are consistently easier to identify than translated movie titles, as reflected by higher scores across all evaluation metrics. We attribute this discrepancy primarily to the fact that movie titles are substantially shorter than advertising slogans, offering fewer tokens and less contextual redundancy, which makes cultural style inference more challenging for LLMs. Advertising slogans, by comparison, provide richer contextual information that facilitates more reliable identification of cultural cues. Human participants, however, can reliably recognize cultural stylistic cues even from very short texts. Second, \textbf{LLMs display systematic, text-type-dependent biases in label assignment}. For movie titles, LLMs tend to misclassify CN-labeled instances as HK, whereas the opposite pattern is observed for advertising slogans. This asymmetry is reflected in recall scores: recall is higher when HK is the primary label in the movie title domain, but this pattern reverses in the slogan domain. Together, these results indicate that LLMs' cultural stylistic biases are not uniform, but vary substantially with the textual genre and informational structure. Regarding training language exposure, \textbf{models primarily trained on Chinese corpora generally outperform those trained mainly on English}, suggesting that access to Chinese-language data is beneficial for recognizing Chinese cultural styles. However, this advantage does not straightforwardly extend to Cantonese. Models trained on more Cantonese data (i.e., SenseNova models) do not consistently yield superior performance on cultural stylistic identification. This limitation may stem from several factors, including the relatively limited scale of available Cantonese corpora, the divergence between written and spoken Cantonese, and a higher level of annotation noise. 

\paragraph{Error Study.}
We conduct an error analysis to better understand the failure modes of LLMs, which reveals three major sources of error at different levels. \textbf{(1) Inherent ambiguity in the data.} A subset of errors stems from cases where cultural stylistics are genuinely difficult to distinguish, even for human annotators. Specifically, 17 instances in \textsc{C\textsuperscript{4}Styli}-T, primarily from films released around 2025, exhibit highly similar CN and HK titles. For example, the CN title of \textit{Mickey 17} is ``\begin{CJK}{UTF8}{gbsn}编号17\end{CJK} (Number 17),'' while the HK title is ``\begin{CJK}{UTF8}{bsmi}米奇17號\end{CJK} (\textit{Mickey 17}).'' Such translations offer minimal stylistic cues, rendering reliable attribution inherently challenging. A similar pattern appears in 30 instances from \textsc{C\textsuperscript{4}Styli}-S advertising slogans. For example, the 2001 HK slogan of Ausupreme, ``\begin{CJK}{UTF8}{bsmi}守護健康，澳至尊\end{CJK} (\textit{lit.} \textit{Safeguarding Your Health, Ausupreme}),'' is stylistically compatible with both CN and HK conventions. These cases, therefore, reflect an upper bound imposed by data ambiguity rather than model-specific limitations.
\textbf{(2) LLM heuristic-based reasoning in movie title classification.}
Beyond data ambiguity, LLMs exhibit systematic failures in classifying movie titles due to stereotypical and oversimplified heuristics. Models tend to reduce stylistic judgment to a single literal--adaptive axis, implicitly associating more literal renderings with CN and freer adaptations with HK. As a result, titles that adopt adaptive yet concise translation strategies are often misclassified. For instance, given the HK title of \textit{Ready Player One}, ``\begin{CJK}{UTF8}{bsmi}挑戰者1號\end{CJK} (\textit{lit.} \textit{Challenger 1}),'' the model incorrectly classified it as CN, despite the title reflecting a narrative-driven translation strategy characteristic of HK conventions. Moreover, such reasoning is highly fragile: even changing the script from Simplified to Traditional Chinese can reverse the model’s judgment. For the HK title of \textit{Alpha} (``\begin{CJK}{UTF8}{bsmi}馴狼紀\end{CJK} [\textit{lit.} \textit{Taming the Wolf}]''), the same model interpreted the title as a creative HK-style adaptation under a Simplified Chinese prompt, but as a literal translation under a Traditional Chinese prompt.
\textbf{(3) LLMs' limited understanding of HK-style humor and rhetoric in advertising slogans.}
A distinct failure pattern is observed in advertising slogans, where LLMs struggle to recognize culturally grounded humor and rhetorical strategies specific to HK. Although models often state that HK slogans emphasize interaction, wordplay, and playfulness, this knowledge remains largely superficial. A representative failure is the 2025 HK advertisement for the Value Partners Asian Income Fund: ``\begin{CJK}{UTF8}{bsmi}進退之間，如何攻守兼利？\end{CJK} (\textit{lit. Between advancing and retreating, how can one achieve both growth and protection?}).'' The slogan employs metaphorical and subtly humorous rhetoric, drawing on strategic and military imagery to frame personal finance decisions, a recurring pattern in HK advertising discourse. Nevertheless, all evaluated LLMs misclassified such cases as CN, indicating that their judgments are driven by a stereotype that CN advertising favors abstract and grand narratives, instead of sensitivity to culturally specific rhetorical strategies.

\section{Can LLMs Produce Cultural Stylistics?}
\paragraph{Method.}
Productive capability is another crucial aspect of assessing LLMs' cultural awareness. Thus, we formulate a conditional text generation task in which LLMs are instructed to produce texts tailored to a target region. For movie titles, the LLM is provided with the original English title and a plot summary in English, and is instructed to generate a Chinese title that reflects the stylistic conventions of either HK or CN. For advertising slogans, the LLM receives the brand name, product information, and release year, and is asked to generate a slogan appropriate for the specified region.

\begin{figure}[tb]
\centering
\resizebox{\linewidth}{!}{\begin{tabular}{cccccc}
    \toprule
    \multicolumn{2}{c}{\multirow{2}{*}{\textbf{Model}}} & \multicolumn{2}{c}{\textbf{\textsc{C\textsuperscript{4}Styli}-T}} & \multicolumn{2}{c}{\textbf{\textsc{C\textsuperscript{4}Styli}-S}} \\
    \cmidrule(lr){3-4} \cmidrule(lr){5-6}
    & & \textbf{HK (\%)} & \textbf{CN (\%)} & \textbf{HK (\%)} & \textbf{CN (\%)}\\
    \midrule
    \multirow{7}{*}{En}&  o4-mini & 82.9  & \textbf{81.7} & 55.5  & 58.9 \\
    & GPT-4.1 & \textbf{93.6}  & 73.2 & 59.6  & 61.0 \\
    & Llama-3-8B-Instruct & 74.3  & 36.5 & 44.8  & 67.5 \\
    & Llama-3-70B-Instruct & 78.6  & 44.1 & 43.2  & \textbf{69.1} \\
    & gemma-3-4b-it & 82.8  & 21.5 & 63.4  & 61.7 \\
    & gemma-3-12b-it & 79.6  & 72.2 & 58.6  & 60.2 \\
    & gemma-3-27b-it & 81.3  & 68.9 & \textbf{70.2}  & 62.1 \\
    \midrule
    \multirow{9}{*}{Zh} & DeepSeek-V3.2 & \textbf{90.4}  & 77.6 & \textbf{64.9}  & 63.0 \\
    & DeepSeek-V3.2 (Thinking) & 88.6  & \textbf{81.4} & 53.6  & \textbf{64.3} \\
    & Qwen2.5-7B-Instruct & 66.5  & 63.3 & 56.3  & 60.6 \\
    & Qwen2.5-14B-Instruct & 85.8  & 54.7 & 46.5  & 64.1 \\
    & Qwen2.5-32B-Instruct & 84.9  & 76.1 & 48.9  & 64.1 \\
    & Qwen2.5-72B-Instruct & 84.4  & 79.3 & 52.5  & 61.0 \\
    & Qwen3-8B & 75.0  & 72.1 & 42.8  & \textbf{63.3} \\
    & Qwen3-14B & 76.3  & 78.9 & 48.2  & 61.4 \\
    & Qwen3-32B & 83.1  & 69.2 & 49.0  & 60.9 \\
    \midrule
    \multirow{2}{*}{Yue} & SenseChat-5-Cantonese & 70.2  & \textbf{57.6} & 46.8  & 60.0 \\
    & SenseNova-V6-5-Turbo & \textbf{83.1} & 57.1 & \textbf{50.2}  & \textbf{62.1} \\
    \bottomrule
    \end{tabular}
}
\caption{LLM performance in cultural stylistic generation, automatically evaluated by classifiers.}
\label{fig:gen_eval_auto}
\end{figure}

\begin{figure}[t]
    \centering
    \includegraphics[width=\linewidth]{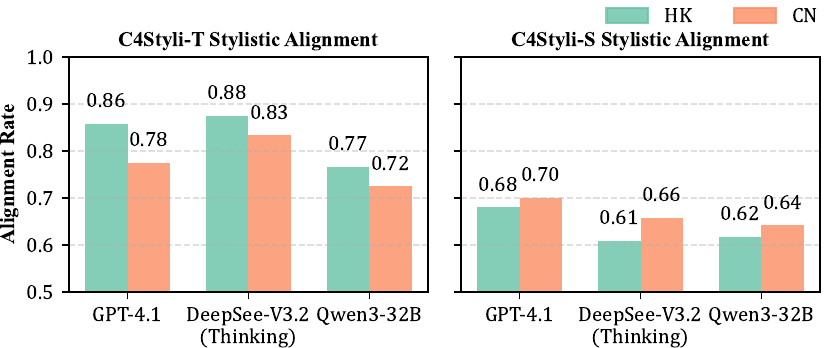}
    \caption{LLM performance in cultural stylistic generation, evaluated by human annotators. Krippendorff's $\alpha$ is 0.776, indicating a substantial agreement.}
\label{fig:gen_eval_human}
\end{figure}

\paragraph{Evaluation.}
For \textbf{automatic evaluation} of the cultural stylistics of LLM generations, we train lightweight \textbf{classifiers} on frozen hidden representations extracted from Qwen3-8B. We concatenate the hidden states from the last 3\textsuperscript{rd} and 4\textsuperscript{th} layers and apply mean pooling over tokens to obtain fixed-length embeddings. Separate MLP classifiers are trained for movie titles and advertising slogans. The classifier achieves an accuracy of 82.13\% on the movie title generation task and 89.73\% on the advertising slogan generation task. We use these classifiers to determine the \textit{proportion of generated texts that align with the target cultural style}.
In addition, we conducted a \textbf{human evaluation} on outputs generated by \textsc{\textit{GPT-4.1}}, \textsc{\textit{Qwen3-32B-Instruct}}, and \textsc{\textit{DeepSeek-V3.2 (thinking)}}. For each instance, annotators were presented with the input context alongside the model-generated text. They were asked to judge whether the text more closely reflected HK culture, CN culture, or whether the style was not apparent. Each instance was independently annotated by two annotators, one from HK and one from CN. In cases of disagreement, a third annotator was consulted to determine the final label. For each model and each domain, we randomly sampled 240 instances.

\begin{table}[tb]
\centering
\scriptsize
\begin{tabularx}{\linewidth}{>{\centering\arraybackslash}m{2cm} X} 
\toprule
\cellcolor{gray!15} & \cellcolor{gray!15} Movie Title: Materialists \\
\cellcolor{gray!15} & \cellcolor{gray!15} Plot: A young, ambitious New York City matchmaker finds herself torn between the perfect match and her imperfect ex. \\
\cellcolor{gray!15} \multirow{-5}{*}{\textbf{Movie Context}} & \cellcolor{gray!15} Year: 2025 \\

\midrule
\multirow{2}{*}{\makecell[c]{DeepSeek-V3.2\\(Thinking)}} 
& CN-Style: \begin{CJK}{UTF8}{gbsn}物质主义者\end{CJK} (Materialists) \\
& HK-Style: \begin{CJK}{UTF8}{bsmi}紐約愛情選擇題\end{CJK} (A New York Love Dilemma)\\
\midrule
\cellcolor{gray!15} & \cellcolor{gray!15} Brand Name: \begin{CJK}{UTF8}{gbsn}中国人寿\end{CJK} (China Life)  \\
\multirow{-2}{*}{\cellcolor{gray!15}\textbf{Slogan Context}} & \cellcolor{gray!15} \begin{CJK}{UTF8}{gbsn}Product: 保险 (Insurance)\end{CJK} \\
\midrule
\multirow{4}{*}{\makecell[c]{DeepSeek-V3.2\\(Thinking)}} 
& CN-Style: \begin{CJK}{UTF8}{gbsn}中国人寿，伴您一生平安。 \end{CJK} (China Life, safeguarding your peace for a lifetime.) \\
& HK-Style: \begin{CJK}{UTF8}{bsmi}中國人壽，守護香港每一家。\end{CJK} (China Life, protecting every family in Hong Kong.) \\
\bottomrule
\end{tabularx}
\caption{LLM-generated texts in different cultural styles.}
\label{tab:cases}
\end{table}

\paragraph{Overall Performance.}
\Cref{fig:gen_eval_auto} and \Cref{fig:gen_eval_human} presents LLM performance on cultural stylistic generation for translated movie titles and advertising slogans. Here, HK and CN scores indicate how well the generated text conforms to the target HK or CN cultural style, respectively. 
Across all models, \textbf{generation quality is consistently higher for translated movie titles than for advertising slogans}. For titles, many models achieve accuracies above 80\% when generating HK-style text, whereas slogan generation remains substantially more challenging, with most accuracies ranging between 45\% and 65\%. This pattern contrasts with the models' discrimination performance. This phenomenon is intuitive: translated movie titles are typically very short, which makes cultural distinctions harder to recognize but easier to generate, whereas advertising slogans tend to be longer and stylistically richer, making them easier to discriminate but more challenging to generate. 
\textbf{A notable asymmetry between HK and CN stylistic generation emerges across domains.} For translated movie titles, models tend to achieve higher accuracy when generating HK-style text than CN-style text, whereas for advertising slogans, CN-style generation often attains higher accuracy. This pattern indicates that LLMs’ generation biases with respect to cultural stylistics are domain-dependent.
\textbf{The effect of the model's primary training language on cultural stylistic generation remains inconclusive.} Performance differences are largely model-specific rather than language-specific, and comparable accuracies can be observed across English-, Chinese-, and Cantonese-oriented models. 

\paragraph{Case Study.}
We present representative examples generated by DeepSeek-V3.2 (Thinking) in \Cref{tab:cases}. The results demonstrate that the model effectively captures the essence of CN-style literalism for movie titles. In contrast, for HK-style adaptations, the model provides localized titles that align more closely with the underlying narrative or plot. Regarding slogans, while the model successfully reflects the concise and abstract nature of the CN-style, it struggles to replicate the distinctive humor and witty charm characteristic of Hong Kong’s creative style, indicating a persistent gap in capturing nuanced cultural flair.
Furthermore, a notable limitation is the model’s tendency to rely on a superficial and somewhat tautological approach to localization. It frequently resorts to the explicit inclusion of the word ``\begin{CJK}{UTF8}{bsmi}香港\end{CJK} (Hong Kong)'' as a lexical crutch to signal regional identity, rather than naturally embedding cultural or linguistic nuances. This rigid dependency on explicit geographic markers highlights a significant deficit in LLMs' genuine understanding of cultural stylistics.

\section{Do LLMs Internally Encode Cultural Stylistics?}

\begin{figure}[t]
    \centering
    \includegraphics[width=0.85\linewidth]{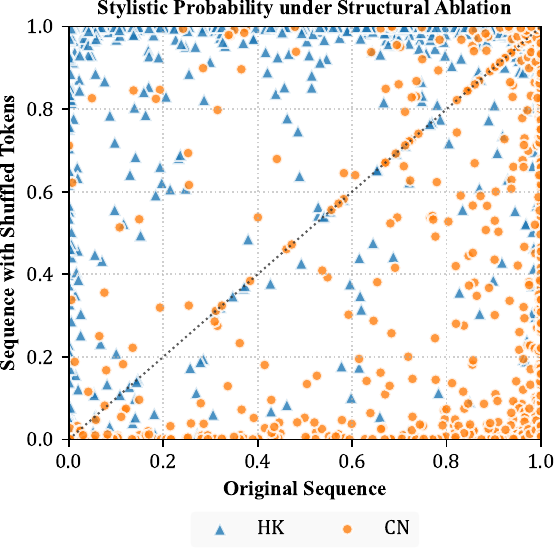}
    \caption{Stylistic Probabilities: Original Text vs. Sequence with Shuffled Tokens. The dotted line is $y=x$ for reference.}
    \label{fig:probe}
\end{figure}

\paragraph{Task.} 
To discern whether LLMs leverage integrated structural understanding or merely rely on superficial lexical anchors, we evaluate representational depth by disrupting syntactic structures. We employ a Logistic Regression (LR) probe, trained on the frozen hidden states of Qwen3-8B to classify cultural styles (CN vs. HK). For each sample in the val set, we utilize the LR probe to obtain the probability of the ground-truth label. By comparing these scores across original sequences and their shuffled variants (which preserve token frequency but eliminate syntax), we quantify the marginal contribution of structural information to cultural identity. This examines whether the model's stylistic discernment is rooted in high-level syntax or isolated token features.

\paragraph{Results.}
\Cref{fig:probe} illustrates the classification probabilities of sequences before and after token shuffling for each instance. The structural ablation results yield a critical finding: \textbf{the LLM fails to achieve a genuine understanding of HK culture, particularly regarding its aesthetic stylistics}. For HK-style texts (blue triangles), the probabilities exhibit a paradoxical increase or remain heavily clustered near $y=1.0$ following the disruption of syntactic structure. This suggests that the model's recognition of HK culture is primarily driven by superficial lexical anchors rather than holistic structural understanding. Conversely, for CN-style texts (orange circles), the probabilities undergo a catastrophic drop upon shuffling, with many samples shifting from $x=1.0$ to $y \approx 0.0$. This sensitivity to structural integrity demonstrates that the identification of CN-style identity relies significantly on integrated syntactic coherence.

\section{Conclusion}
This work studies cultural awareness in LLMs through \emph{aesthetic stylistics}, an implicit cultural dimension governing how meaning is expressed. We first introduce \textsc{C\textsuperscript{4}Styli}, a cross-cultural Chinese--Chinese benchmark built on highly stylized texts.
With this dataset, our experiments yield three findings. First, LLMs significantly underperform humans in recognizing cultural stylistics, with strong domain sensitivity. Second, stylistic recognition and production are decoupled: accurate identification does not imply culturally aligned generation. Third, representation analyses reveal an encoding asymmetry: CN stylistics are captured in more distributed structures, whereas HK stylistics rely on sparse, high-salience lexical cues, reflecting superficial cultural awareness.

\section*{Ethical Statement}
\paragraph{Dataset Collection.} This work uses only publicly available data sources collected from open-access websites and publicly released materials. The data do not contain private, sensitive, or personally identifiable information, and are used strictly in accordance with their original usage terms and for research purposes only.

\paragraph{LLM Evaluation.} During LLM-based evaluation, no personal or sensitive information was provided as input to any model. The prompts and test instances consist solely of publicly available or model-generated text.

\paragraph{Human Annotation.} For human evaluation, we recruited annotators with relevant cultural and linguistic backgrounds. Participation was voluntary, and annotators were informed of the purpose of the study. The annotation tasks involved labeling model-generated content and did not expose annotators to harmful or sensitive material. No personal information about the annotators was collected or disclosed.

\paragraph{Use of Models.} Finally, while we release trained models and code to support reproducibility, we encourage their responsible and appropriate use in accordance with relevant ethical guidelines.

\section*{Acknowledgments}
We thank the anonymous area chair and anonymous reviewers for their insightful comments and valuable feedback during the review process. This study is funded by the Research Grants Council (project code: T43-518/24-N and PolyU/15213323) under the University Grants Committee, Hong Kong Special Administrative Region Government.


\bibliographystyle{named}
\bibliography{ijcai26}

\appendix
\section{Experimental Details}
\subsection{Accessed Models}
For behavioral recognition and productive competence, we evaluate the cultural awareness of 18 large language models (LLMs), spanning six major open-source and closed-source model families. The evaluated models were primarily trained on English, Simplified Chinese, and Cantonese corpora. We group the models into the following categories: (1) OpenAI series, including {o4-mini} and {GPT-4.1}; (2) Meta-Llama series, including {Meta-Llama-3-8B-Instruct} and {Meta-Llama-3-70B-Instruct}; (3) Google Gemma series, including {Gemma-3-4B-it}, {Gemma-3-12B-it}, and {Gemma-3-27B-it}; (4) DeepSeek series, including {DeepSeek-V3.2} in both non-thinking and thinking modes; (5) Qwen series, including {Qwen2.5-7B-Instruct}, {Qwen2.5-14B-Instruct}, {Qwen2.5-32B-Instruct}, {Qwen2.5-72B-Instruct}, {Qwen3-8B}, {Qwen3-14B}, and {Qwen3-32B}; and (6) SenseNova series, including {SenseChat-5-Cantonese} and {SenseNova-V6-5-Turbo}.

When available, we report the official knowledge cutoff dates disclosed by model providers. For most evaluated models, however, exact knowledge cutoff information is not publicly specified; in such cases, we report the model release or announcement date for reference. Specifically, {o4-mini} and {GPT-4.1} were released in Jun 2024;\footnote{\url{https://platform.openai.com/docs/models/o3}\quad\url{https://openai.com/index/gpt-4-1/}} {DeepSeek-V3.2} was released in Jan 2024; Meta-Llama-3 models' knowledge cutoff dates were Mar 2023 (8B) and Dec 2023 (70B), respectively;\footnote{\url{https://github.com/meta-llama/llama-models/blob/main/models/llama3/MODEL_CARD.md}} Gemma-3 models' knowledge cutoff date was in Aug 2024;\footnote{\url{https://ai.google.dev/gemma/docs/core/model_card_3}} and the Qwen2.5 series was announced in September 2024,\footnote{\url{https://qwenlm.github.io/blog/qwen2.5/}} with the Qwen3 series released Apr 2025.\footnote{\url{https://qwenlm.github.io/blog/qwen3/}} {SenseChat-5-Cantonese} was released on 23 Apr 2024; and {SenseNova-V6-5-Turbo} was released in Jun 2025.\footnote{\url{https://console.sensecore.cn/cn-sh-01/help/docs/model-as-a-service/nova/release}}

\subsection{Implementation Settings}
We conduct experiments on our collected dataset, \textsc{C\textsuperscript{4}Styli}. Closed-source models are evaluated via their respective APIs, while open-source models are run using vLLM on 4$\times$NVIDIA A6000 GPUs (48GB each). Unless stated otherwise, the sampling temperature is set to 0.7. Probes are trained on a single NVIDIA A6000 GPU.

\subsection{Prompts in Experiments}
\paragraph{Prompt Designs.}
The prompts used to test the LLMs' discrimination capacities are shown as follows:
\begin{tcolorbox}[
    enhanced,
    colback=white, colframe=black!20, 
    boxrule=0.2mm,
    arc=0.5mm, 
    fontupper=\ttfamily\scriptsize,
    fonttitle=\scriptsize\bfseries,
    breakable
]
\textbf{\underline{Simplified Chinese slogan classification prompt:}}\\\\
\begin{CJK}{UTF8}{gbsn}请分析以下广告标语的文化风格：
\\\\
品牌名称: \{company\}\{product\_info\}\{year\_info\}
\\\\
广告标语: \{slogan\}
\\\\
请根据广告标语在语言风格、表达方式与文化取向等方面的特征，
判断其更接近「大陆广告风格」还是「香港广告风格」。
\\\\
请综合以下维度进行分析（不要求每条都出现，但需整体权衡）：
\\\\
1. 表达策略与创意取向
    - 大陆广告标语：倾向于结构完整、逻辑清晰的表达方式，强调广告信息的可理解性与整体表达的规范性，创意多服务于清晰传达主题。
    - 香港广告标语：更强调创意表达本身，常通过重构语言、制造意外感来吸引注意力，创意手法灵活，形式感强。
\\\\
2. 语言风格与修辞手段
    - 大陆广告标语：语言偏正式、书面化，常使用成语、四字格、排比等修辞结构，语气较为庄重稳定。
    - 香港广告标语：语言口语化程度高，常使用粤语表达、中英混合及双关语，风格轻松活泼，富有节奏感与趣味性。
\\\\
3. 叙事方式与内容呈现
    - 大陆广告标语：倾向于概括式或概念化表达，强调整体意义与信息传达，叙事层次相对抽象。
    - 香港广告标语：更偏向具体情境或生活化叙事，通过细节、场景或人物引发共鸣，叙事方式更具画面感。
\\\\
4. 传播取向与受众互动方式
    - 大陆广告标语：以"说明—引导"为主要功能，强调信息的完整性和规范性，受众角色相对被动。
    - 香港广告标语：以"吸引—参与"为导向，通过语言趣味性与文化共鸣提升记忆度与传播性。
\\\\
请基于以上特征进行推理判断。
\\\\
输出格式要求：
\\\\
请仅以 JSON 格式输出，包含以下字段：
\\\\
- "slogan": 广告标语原文
\\\\
- "is\_mainland": 是否更接近大陆广告风格（true / false）
\\\\
- "is\_hongkong": 是否更接近香港广告风格（true / false）
\\\\
- "confidence": 判断置信度（0–1 之间的小数）
\\\\
- "reasoning": 判断依据说明（需明确对应上述分析维度）
\end{CJK}
\end{tcolorbox}

\begin{tcolorbox}[
    enhanced,
    colback=white, colframe=black!20, 
    boxrule=0.2mm,
    arc=0.5mm, 
    fontupper=\ttfamily\scriptsize,
    fonttitle=\scriptsize\bfseries,
    breakable
]
\textbf{\underline{Traditional Chinese slogan classification prompt:}}\\\\
\begin{CJK}{UTF8}{bsmi}請分析以下廣告標語的文化風格：
\\\\
品牌名稱: \{brand\}\{product\_info\}\{year\_info\}
\\\\
廣告標語: \{slogan\}
\\\\
請根據廣告標語在語言風格、表達方式與文化取向等方面的特徵，
判斷其更接近「大陸廣告風格」還是「香港廣告風格」。
\\\\
請綜合以下維度進行分析（不要求每條都出現，但需整體權衡）：
\\\\
1. 表達策略與創意取向
    - 大陸廣告標語：傾向於結構完整、邏輯清晰的表達方式，強調廣告信息的可理解性與整體表達的規範性，創意多服務於清晰傳達主題。
    - 香港廣告標語：更強調創意表達本身，常通過重構語言、製造意外感來吸引注意力，創意手法靈活，形式感強。
\\\\
2. 語言風格與修辭手段
    - 大陸廣告標語：語言偏正式、書面化，常使用成語、四字格、排比等修辭結構，語氣較為莊重穩定。
    - 香港廣告標語：語言口語化程度高，常使用粵語表達、中英混合及雙關語，風格輕鬆活潑，富有節奏感與趣味性。
\\\\
3. 敘事方式與內容呈現
    - 大陸廣告標語：傾向於概括式或概念化表達，強調整體意義與信息傳達，敘事層次相對抽象。
    - 香港廣告標語：更偏向具體情境或生活化敘事，通過細節、場景或人物引發共鳴，敘事方式更具畫面感。
\\\\
4. 傳播取向與受眾互動方式
    - 大陸廣告標語：以"說明—引導"為主要功能，強調信息的完整性和規範性，受眾角色相對被動。
    - 香港廣告標語：以"吸引—參與"為導向，通過語言趣味性與文化共鳴提升記憶度與傳播性。
\\\\
請基於以上特徵進行推理判斷。
\\\\
輸出格式要求：
\\\\
請僅以 JSON 格式輸出，包含以下字段：
\\\\
- "slogan": 廣告標語原文
\\\\
- "is\_mainland": 是否更接近大陸廣告風格（true / false）
\\\\
- "is\_hongkong": 是否更接近香港廣告風格（true / false）
\\\\
- "confidence": 判斷置信度（0–1 之間的小數）
\\\\
- "reasoning": 判斷依據說明（需明確對應上述分析維度）
\end{CJK}
\end{tcolorbox}

\begin{tcolorbox}[
    enhanced,
    colback=white, colframe=black!20, 
    boxrule=0.2mm,
    arc=0.5mm, 
    fontupper=\ttfamily\scriptsize,
    fonttitle=\scriptsize\bfseries,
    breakable
]
\textbf{\underline{Simplified Chinese title classification prompt:}}\\\\
\begin{CJK}{UTF8}{gbsn}请分析以下电影标题的翻译风格：
\\\\
英文标题: \{title\_en\}
\\\\
英文简介:\{summary\_en\}
\\\\
中文简介: \{summary\}
\\\\
中文标题: \{title\}
\\\\
根据中文简介和中文标题，请判断这个中文标题更接近大陆译名还是香港译名。
请勿根据简体或繁体字形直接判断，简繁体信息已被**刻意扰乱**。
\\\\
请结合以下翻译特征进行综合分析（不要求每条都出现，但需整体权衡）：
\\\\
1. 翻译策略取向
    - 大陆译名：更倾向于逐字或近似逐字翻译，较多保留英文原名的表达形式
    - 香港译名：常对原片名进行较大幅度改写，仅保留部分语境或核心内容，语篇重写特征明显
\\\\
2. 译名与影片内容的关系
    - 大陆译名：译名更多对应原文片名本身
    - 香港译名：译名更服务于影片内容、情节或氛围，而不拘泥于原文片名形式
\\\\
3. 语言风格与表达方式
    - 大陆译名：相对正式、书面化，具有一定煽情或情感引导功能
    - 香港译名：更口语化、更贴近观众日常语言，标题表现力强
\\\\
4. 整体受众导向
    - 大陆译名：以信息传达和情感影响为主
    - 香港译名：以吸引观众注意、制造画面感或戏剧张力为主
\\\\
请基于以上特征进行推理判断。
\\\\
请以 JSON 格式输出结果，包含以下字段：
\\\\
- "title": 标题名称
\\\\
- "is\_mainland": 是否为大陆翻译（布尔值）
\\\\
- "is\_hongkong": 是否为香港翻译（布尔值）
\\\\
- "confidence": 置信度（0-1之间的小数）
\\\\
- "reasoning": 判断理由（需明确引用上述特征）
\\\\
只输出 JSON，不要输出其他内容。
\end{CJK}
\end{tcolorbox}

\begin{tcolorbox}[
    enhanced,
    colback=white, colframe=black!20, 
    boxrule=0.2mm,
    arc=0.5mm, 
    fontupper=\ttfamily\scriptsize,
    fonttitle=\scriptsize\bfseries,
    breakable
]
\textbf{\underline{Traditional Chinese title classification prompt:}}\\\\
\begin{CJK}{UTF8}{bsmi}請分析以下電影標題的翻譯風格：
\\\\
英文標題: \{title\_en\}
\\\\
英文簡介: \{summary\_en\}
\\\\
中文簡介: \{summary\}
\\\\
中文標題: \{title\}
\\\\
根據中文簡介和中文標題，請判斷這個中文標題更接近大陸譯名還是香港譯名。
請勿根據簡體或繁體字形直接判斷，簡繁體信息已被**刻意擾亂**。
\\\\
請結合以下翻譯特徵進行綜合分析（不要求每條都出現，但需整體權衡）：
\\\\
1. 翻譯策略取向
    - 大陸譯名：更傾向於逐字或近似逐字翻譯，較多保留英文原名的表達形式
    - 香港譯名：常對原片名進行較大幅度改寫，僅保留部分語境或核心內容，語篇重寫特徵明顯
\\\\
2. 譯名與影片內容的關係
    - 大陸譯名：譯名更多對應原文片名本身
    - 香港譯名：譯名更服務於影片內容、情節或氛圍，而不拘泥於原文片名形式
\\\\
3. 語言風格與表達方式
    - 大陸譯名：相對正式、書面化，具有一定煽情或情感引導功能
    - 香港譯名：更口語化、更貼近觀眾日常語言，標題表現力強
\\\\
4. 整體受眾導向
    - 大陸譯名：以信息傳達和情感影響為主
    - 香港譯名：以吸引觀眾注意、製造畫面感或戲劇張力為主
\\\\
請基於以上特徵進行推理判斷。
\\\\
請以 JSON 格式輸出結果，包含以下字段：
\\\\
- "title": 標題名稱
\\\\
- "is\_mainland": 是否為大陸翻譯（布爾值）
\\\\
- "is\_hongkong": 是否為香港翻譯（布爾值）
\\\\
- "confidence": 置信度（0-1之間的小數）
\\\\
- "reasoning": 判斷理由（需明確引用上述特徵）
\\\\

只輸出 JSON，不要輸出其他內容。
\end{CJK}
\end{tcolorbox}

Then, for each prompt, we randomly select three instances from the training set to serve as in-context exemplars. For the generation task, we adopt a similar prompt structure, except that the translated Chinese movie titles and slogans are omitted and only the target region is provided. However, we observe that when exemplars are included, some LLMs rely heavily on the provided texts, which obscures an assessment of their genuine generative abilities. Therefore, we choose not to include exemplars in the prompts for the generation task.

\begin{figure}[tb]
    \centering
    \begin{subfigure}[t]{\linewidth}
        \centering
        \includegraphics[width=\linewidth]{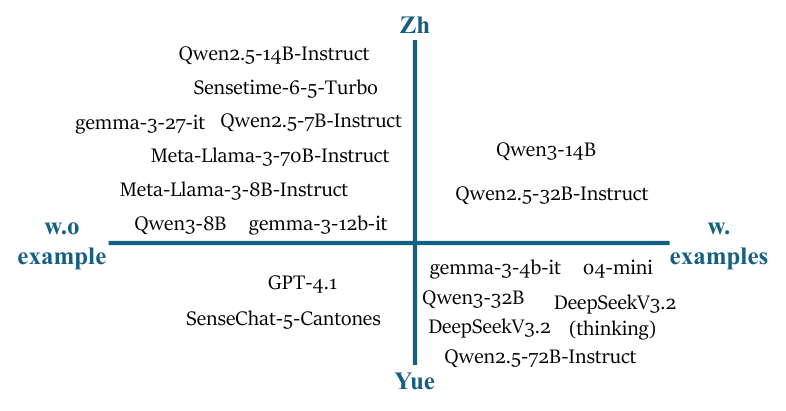}
        \caption{\textsc{C\textsuperscript{4}Styli}-T}
    \end{subfigure}
    \hfill
    \begin{subfigure}[t]{\linewidth}
        \centering
        \includegraphics[width=\linewidth]{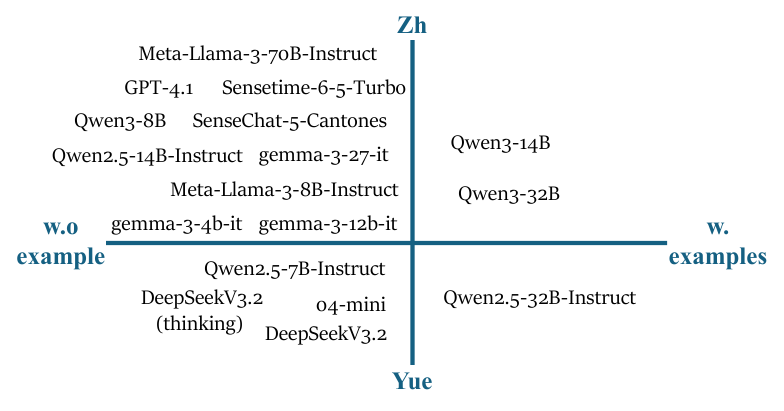}
        \caption{\textsc{C\textsuperscript{4}Styli}-S}
    \end{subfigure}
    \caption{Sensitivity of LLMs to prompt language and the presence of in-context examples.}
    \label{fig:prompt}
\end{figure}

\paragraph{Sensitivity to Language \& Exemplars.}
\Cref{fig:prompt} demonstrates that most models achieve their strongest performance when prompts are written in Simplified Chinese and do not include in-context examples. Overall, the performance gap induced by language and exemplar variations is moderate, with an average accuracy difference of 0.039 across models. The largest variation is observed for SenseChat-5-Cantonese on the \textsc{C\textsuperscript{4}Styli}-S benchmark, with a maximum–minimum accuracy gap of 0.110 across prompt configurations. 

\begin{figure*}[!t]
    \centering
    \includegraphics[width=\textwidth]{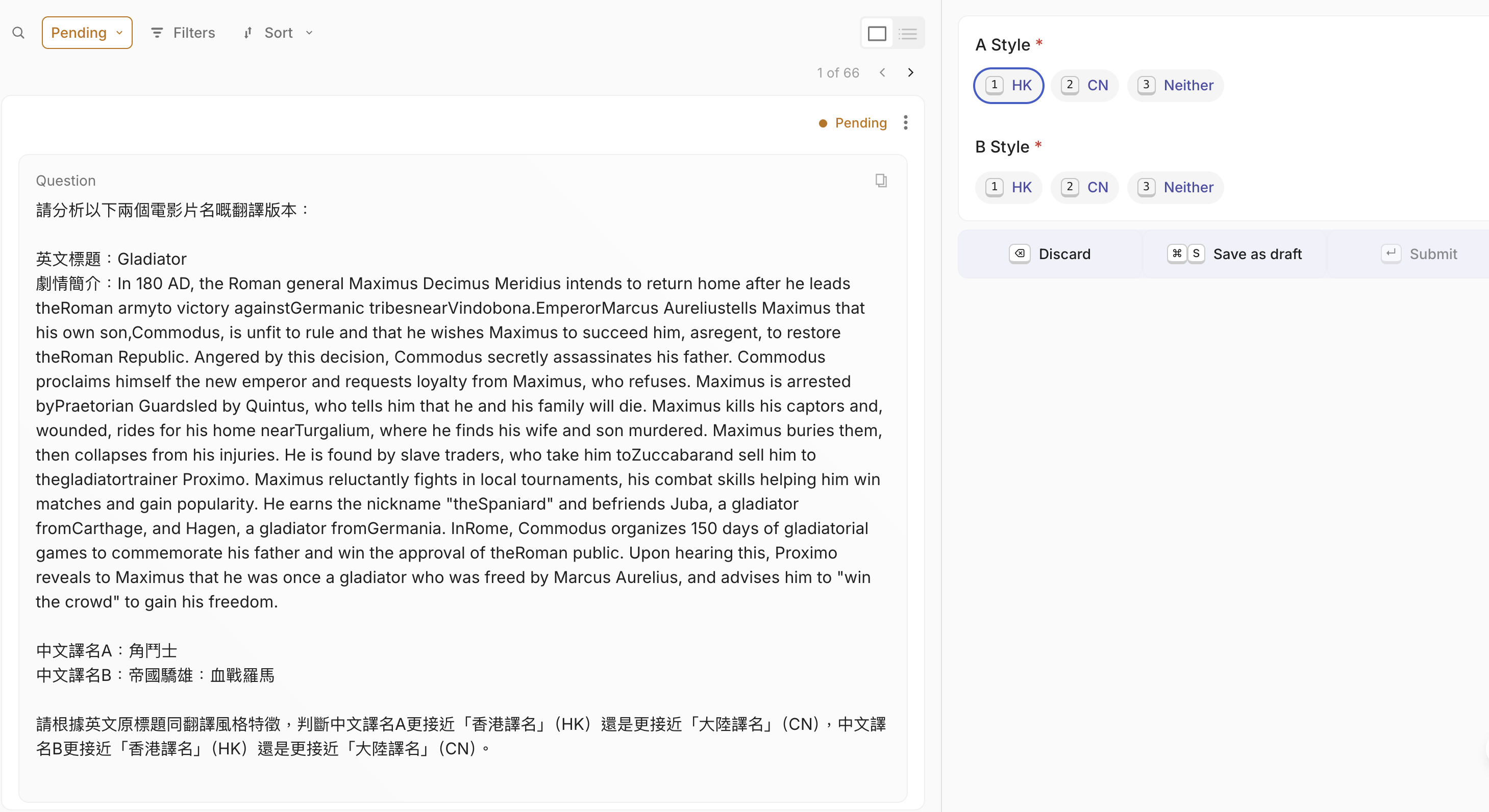}
    \caption{The interface provided to the human participants.}
    \label{fig:appx:ui}
\end{figure*}

\subsection{Human Evaluation}
A total of 12 participants from both Hong Kong and Mainland China were recruited to establish the ground truth for our study. To ensure high-quality labeling, we selected individuals with dual-cultural literacy: six annotators are residents of HK, and six are from CN. All participants demonstrated a deep familiarity with the linguistic and cultural nuances of both regions. 

The annotation was conducted via a web interface designed with HuggingFace and Argilla (\Cref{fig:appx:ui}). Regarding ethical standards, all participants were fully informed of the research objectives and the intended use of the data prior to the commencement of the task. We obtained explicit informed consent from each individual, ensuring they understood that their participation was entirely voluntary. Furthermore, participants were explicitly informed of their right to withdraw from the study at any stage without any penalty or need for justification. To protect participant privacy, all collected data were strictly anonymized.

\end{document}